\documentclass{article}

\usepackage[preprint]{neurips_2023}

\usepackage[utf8]{inputenc} 
\usepackage[T1]{fontenc}    
\usepackage{hyperref}       
\usepackage{url}            
\usepackage{booktabs}       
\usepackage{amsfonts}       
\usepackage{nicefrac}       
\usepackage{microtype}      
\usepackage{xcolor}         
\usepackage{graphicx}
\usepackage{amsmath}
\usepackage{multirow}
\usepackage{color}
\usepackage{subfigure}
\usepackage{pifont}
\usepackage{appendix}
\usepackage[ruled]{algorithm2e}
\usepackage{natbib}
\setcitestyle{numbers,square}

\title{Improving Knowledge Distillation via Transferring Learning Ability}

\author{Long Liu\thanks{Correspondence to: Long Liu<liulong@xaut.edu.cn>},Tong Li,Hui Cheng\\
	School of Automation and Information Engineering\\
	Xi'an University of Technology\\
	No.5 Jinhua South Road, Xi'an, Shaanxi, China \\
}

\begin{document}

\maketitle

\begin{abstract}
Existing knowledge distillation methods generally abide by a teacher-student fashion, where a student neural network solely learns from a well-trained teacher neural network. However, people tend to overlook that, due to the inherent architectural differences, the learning ability of these two neural networks usually varies. The student network performance degrades when the capacity gap between the student and the teacher is too large, and this greatly limits the performance of existing methods. To address this limitation, we propose a novel transferring learning ability knowledge distillation method dubbed SLKD. Specifically, we consider the learning ability of teacher networks as additional knowledge provided by a self-learning teacher. Our approach significantly improves distillation performance by incorporating the learning ability to complement the vanilla KD. We achieve comparable or even better results for image classification tasks on CIFAR-100 and ImageNet datasets. This paper highlights the necessity of transferring the learning ability of teachers, and we hope it will be helpful for future research.
\end{abstract}
\section{Introduction}
\begin{figure}[h]
\center
    \begin{tabular}{cc}
         \includegraphics[width=0.45\textwidth]{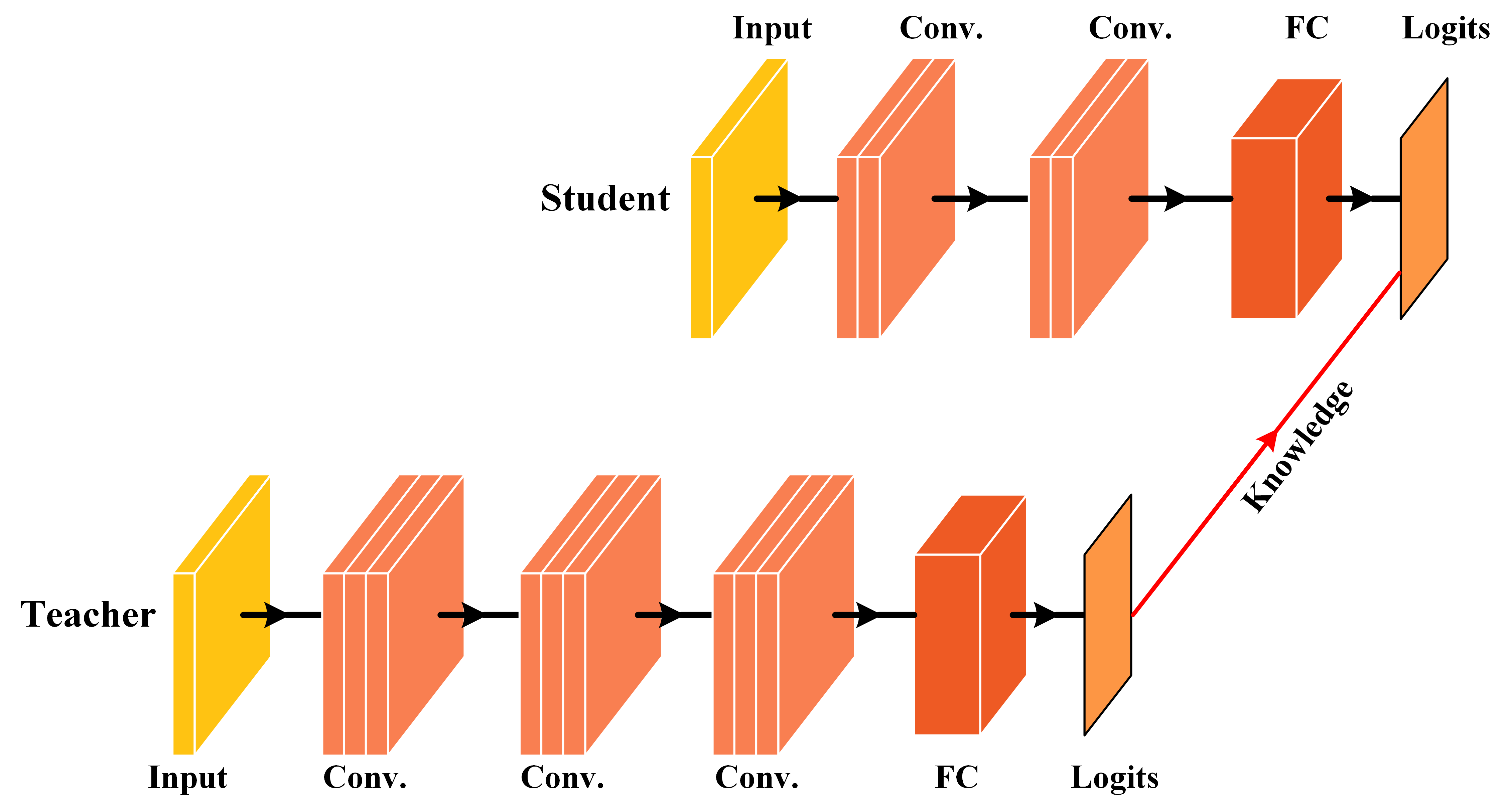}& \includegraphics[width=0.45\textwidth]{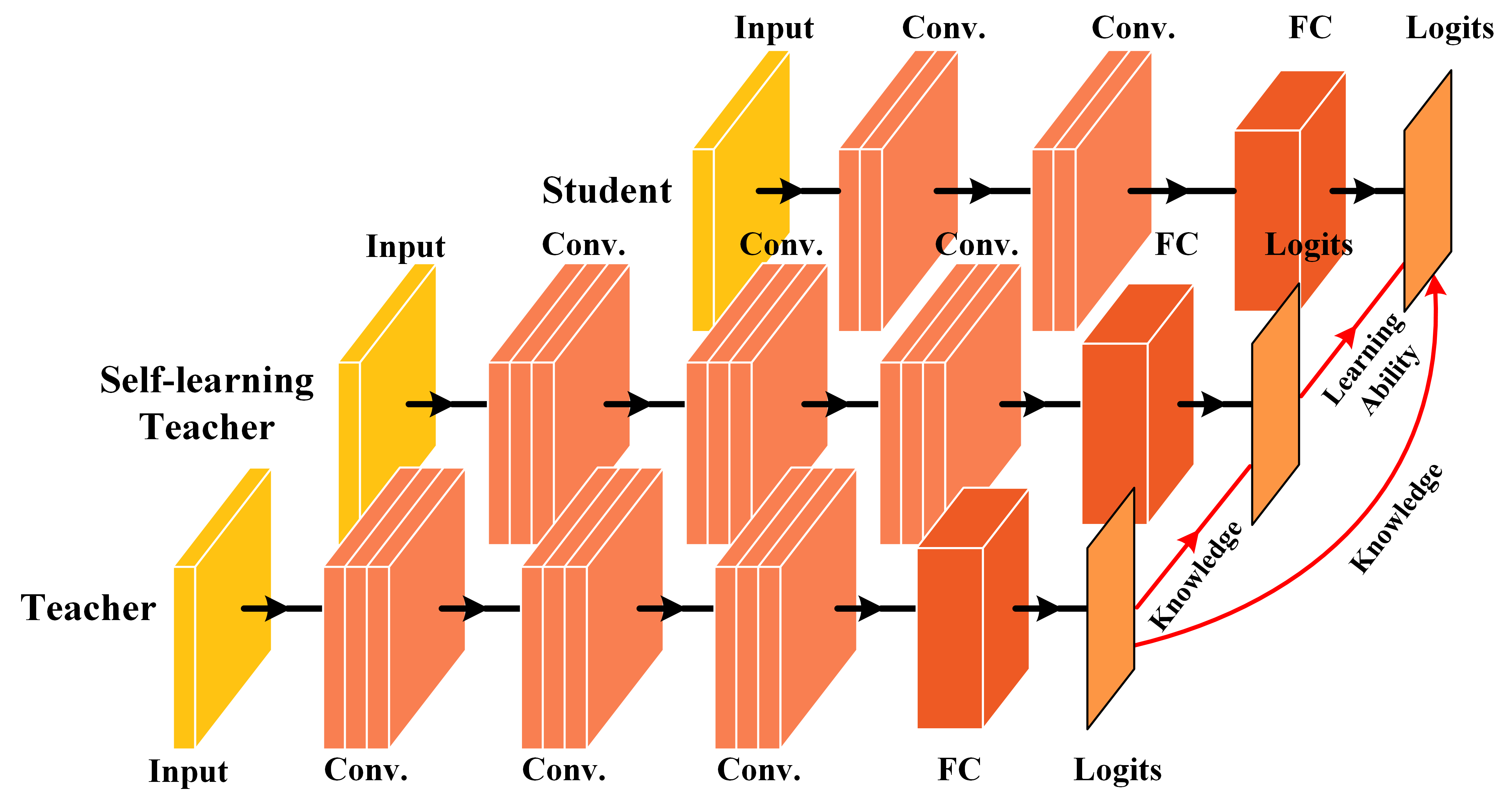} \\
         {(a) Vanilla Knowledge Distillation}&{(b) SLKD}\\ 
    \end{tabular}
    \caption{Comparison of Vanilla KD and our SLKD. The main difference lies in what kind of knowledge is transferred. (a) Vanilla KD transfers the knowledge via KL-Divergence to update the whole student network. (b) Our SLKD measures the KL-Divergence between the student and teacher outputs to transfer the knowledge and also measures the KL-Divergence between the student and SL-T outputs to transfer the learning ability.}
    \label{pic:pic1}
\end{figure}

Deep neural networks have recently exhibited remarkable performance in various computer vision tasks, including image classification\citep{he2016deep,hu2018squeeze,ma2018shufflenet,simonyan2014very}, object detection\citep{he2017piotr,ren2015faster}, and semantic segmentation\citep{long2015fully,zhao2017pyramid}. However, top-performing deep neural networks typically involve numerous parameters, resulting in considerable memory consumption at inference time, making them unsuitable for applications with limited resources. In this regard, knowledge distillation distills the knowledge from one or more well-trained networks(teachers) to boost the performance of less parameterized networks(students).\\
The vanilla knowledge distillation (KD)\citep{hinton2015distilling} minimizes the Kullback-Leibler(KL) divergence between prediction logits of the teacher and student to distill the teacher knowledge(Figure \ref{pic:pic1}a).
FitNet\citep{romero2014fitnets} improves student performance by distilling knowledge from intermediate features. Most knowledge distillation methods have recently focused on distilling knowledge from the intermediate feature maps, achieving remarkable results on various tasks. However, feature-based methods usually involve additional down-sampling operations considering the feature alignment, which can cause information loss and induce additional training costs. In addition, parameters in the fully connected layer are not involved in the procedure of feature-based knowledge distillation, and students cannot acquire the knowledge in the fully connected layer. \\
The prediction logits are generated in the final layer of the deep neural network and have a higher semantic level than deep features. Therefore, logits-based methods should achieve better performance. However, previous works have shown that logits-based methods are inferior to feature-based methods in most cases, especially when the disparity between the teacher and student is significant. These issues can be attributed to the capacity gap between teachers and students. TAKD\citep{mirzadeh2020improved} alleviates the capacity-gap problem by introducing one or more intermediate-sized teacher assistants. DGKD\citep{son2021densely} further improves TAKD by using a densely guided approach to guide the student, thus resolving the error avalanche problem. However, both of these methods require pre-training of teacher assistants, which incurs high training costs. Additionally, the model size and the number of teacher assistants completely depend on prior knowledge, which hinders further generalization of such methods. In knowledge distillation, the capacity gap is reflected by the difference between the output of teachers and students. Students do not have the same learning ability as teachers due to inherent architectural differences, and their output significantly varies from that of the teacher at the initial stage of knowledge distillation, thus causing the poor learning issue of the student network.\\
In this work, we propose a novel distillation framework called Self-Learning Teacher Knowledge Distillation(SLKD) to tackle both the capacity-gap problem and the poor learning issue. Specifically, we introduce a self-learning teacher(SL-T) on top of the vanilla KD(Figure \ref{pic:pic1}b). The SL-T has the same architecture as the teacher but is not pre-trained. During the knowledge distillation, the SL-T network learns from the teacher network as a student while also guiding the student network learning as a teacher. Compared to the teacher, the SL-T output is smoother and more gradual, which benefits the learning procedure. More importantly, since the SL-T is an isomorphic copy of the teacher, its learning trajectory can be approximately equivalent to that of the teacher, which reflects the learning ability of a deep neural network. We can acquire the learning ability of the teacher by tracing the learning trajectory of the SL-T. In this way, we aim to enable the student can acquire a similar learning ability as the teacher, thus mitigating the poor learning issue and the capacity-gap problem. Moreover, to ensure the accuracy and effectiveness of the learning trajectory provided by the SL-T, we design a learning trajectory enhancement strategy inspired by multi-teacher-based knowledge distillation, which fuses the output of multiple SL-Ts. Applying this strategy to knowledge distillation can improve the reliability of the learning trajectory, and therefore further boost the distillation performance.\\
In summary, our contributions are summarized as follows:
\begin{itemize}
\item 
We propose our SLKD solution that introduces an SL-T network to transfer the learning ability of the teacher network, which is accomplished by tracing the learning trajectory of the SL-T network. In this way, we can alleviate the capacity-gap and poor learning issues and improve the knowledge distillation performance by a large margin.
\item
We formulate a general form to transfer the learning ability of a teacher network and theoretically demonstrate its rationality and necessity, which is proved through comparative experiments.
\item 
We achieve state-of-the-art performance on various datasets and network architectures by applying our distillation framework. 
\end{itemize}
\begin{figure}[h]
\centering
\begin{tabular}{cc}
{\includegraphics[width=5cm]{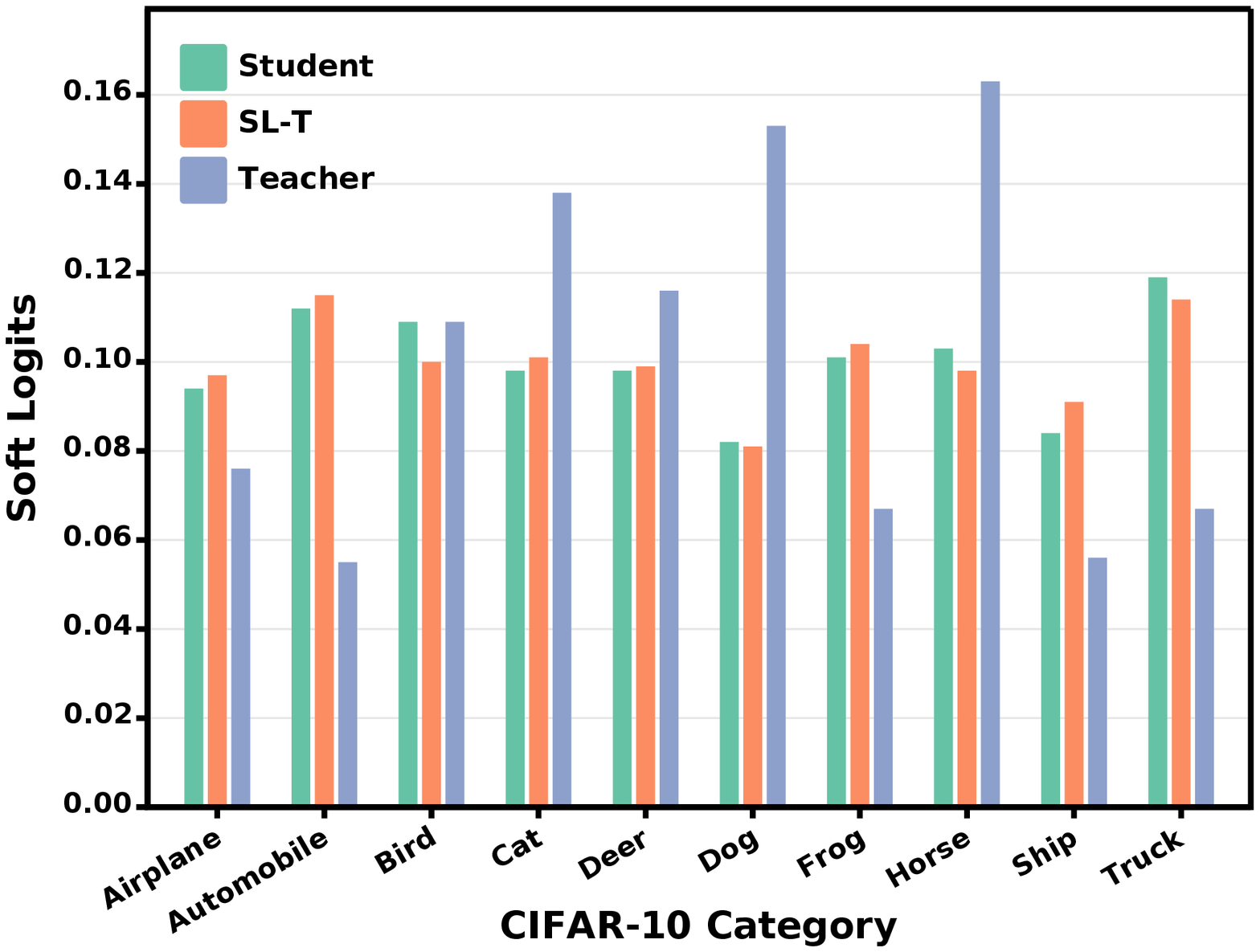}}&{\includegraphics[width=5cm]{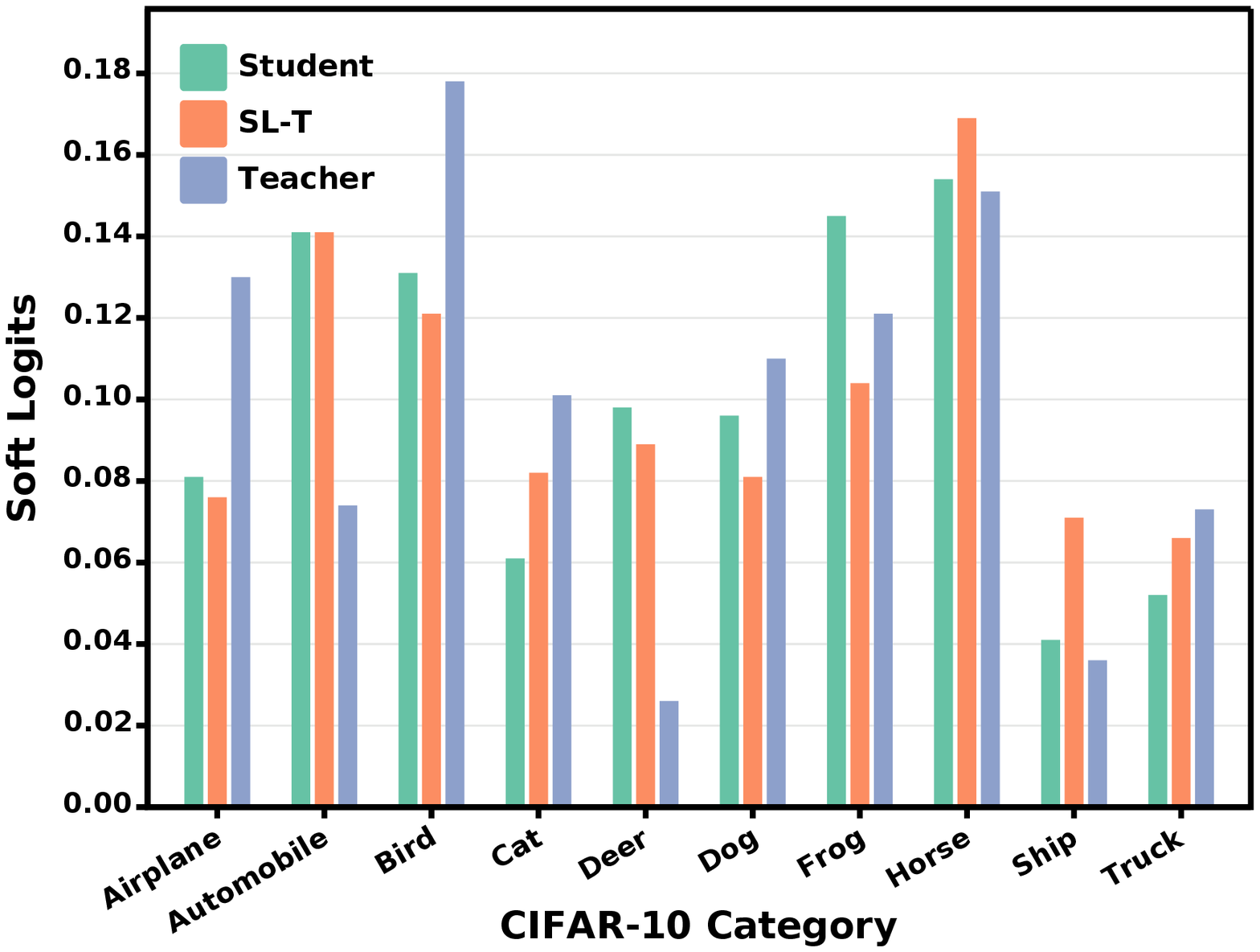}}\\
{(a) VGG13-VGG8}&{(b) ResNet50-MobileNetV2}\\
\end{tabular}
\caption{Comparing the output among the student, SL-T, and teacher network. The KL-Divergence between the student and SL-T is significantly lower than between the student and teacher.}
\label{pic:pic2}
\end{figure}
\section{Related Work}
The original concept of knowledge distillation (KD) was proposed in \citep{hinton2015distilling} to train a smaller student neural network under the supervision of a larger pre-trained teacher. The vanilla KD enables the student to learn the knowledge of a teacher by minimizing the KL-Divergence between prediction logits of the teacher and student and introduces a hyper-parameter temperature to soften the output. FitNet\citep{romero2014fitnets} further improves student performance by minimizing the difference between the student and teacher network deep features of intermediate layers. The following works can be broadly categorized into two types, i.e., logits-based methods\citep{cho2019efficacy,furlanello2018born,yang2019snapshot,zhang2018deep,rezagholizadeh2021pro,zhao2022decoupled} and feature-based methods\citep{heo2019comprehensive,heo2019knowledge,huang2017like,kim2018paraphrasing,park2019relational,peng2019correlation,tian2019contrastive,tung2019similarity,yim2017gift,zagoruyko2016paying,chen2021distilling}.\\
Most of these methods abide by a teacher-student fashion and improve student performance by formulating an external loss function to transfer teacher knowledge. Deep mutual learning (DML)\citep{zhang2018deep} boosts student performance by learning from each other collaboratively without needing a pre-trained teacher, thus reducing training time and computational resources. TAKD\citep{mirzadeh2020improved} alleviates the capacity-gap problem by introducing intermediate-sized teacher assistant(TA) networks. DGKD \citep{son2021densely}further enhances TAKD by densely guiding each TA network with the higher-level TAs as well as the teacher, thus alleviating the error avalanche problem. Pro-KD\citep{rezagholizadeh2021pro} provides a smoother training path for the student by following the teacher training routes, which addresses the capacity-gap and checkpoint search problems. DKD\citep{zhao2022decoupled} reformulates the vanilla KD into target class knowledge distillation and non-target class knowledge distillation, revealing that the vanilla KD loss is a coupled formulation that limits the effectiveness and flexibility of knowledge transfer. In addition, several works also investigate the interpreting of knowledge distillation\citep{cheng2020explaining,phuong2019towards,zhang2022quantifying}.\\
In contrast to existing methods, we suppose that the learning ability can also be viewed as a kind of transferable knowledge to distill. By assimilating the teacher network learning ability, the student network can develop comparable learning competency, which is beneficial for alleviating the capacity-gap problem and poor learning issue, and subsequently improving the knowledge distillation performance.
\begin{figure}[t]
\centering
\includegraphics[width=0.9\textwidth]{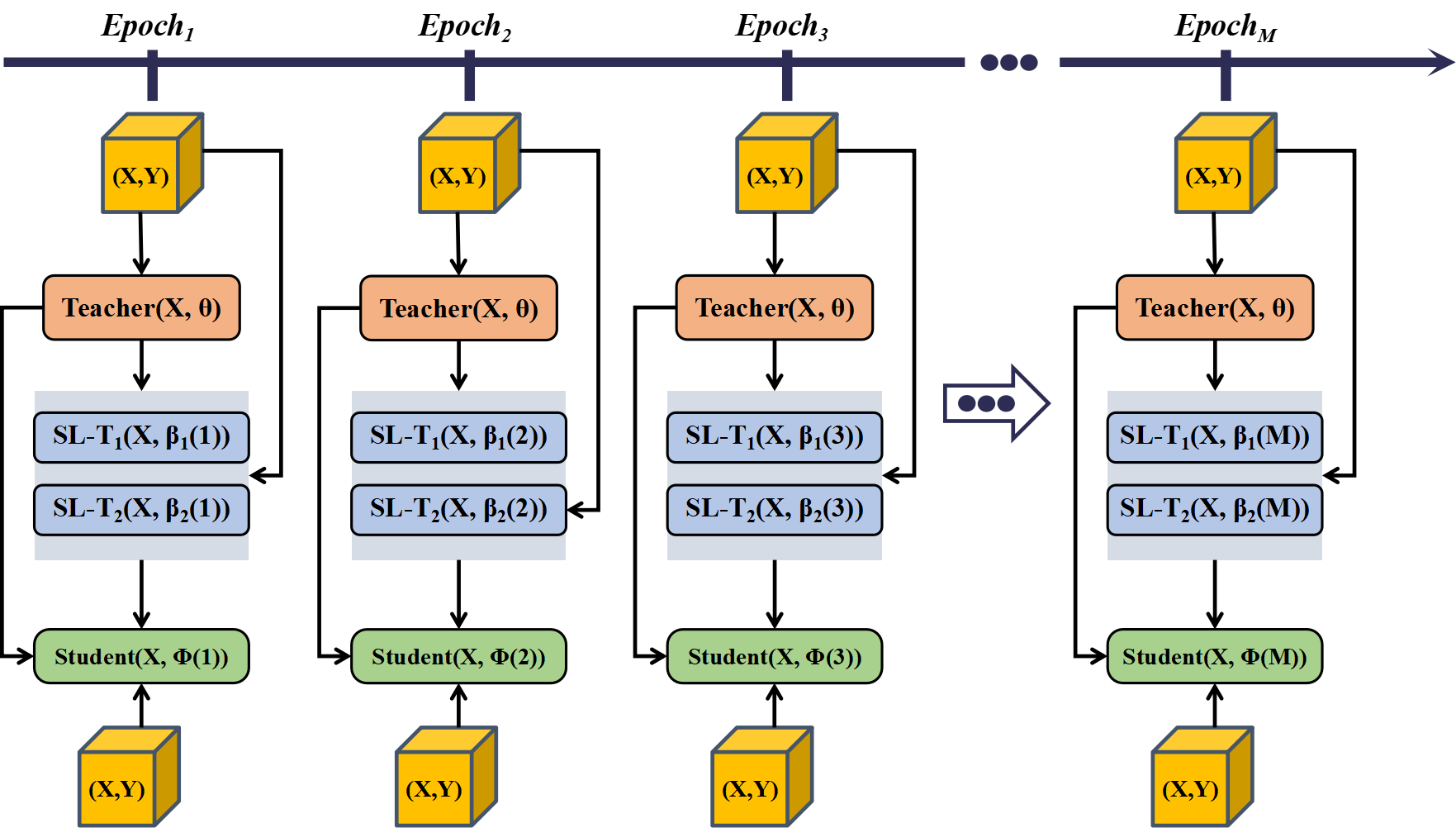}
\caption{An overview of our proposed SLKD. The (X,Y) represents the input, where X is the input sample, and Y is its associated label. When an input sample X is obtained, the student network, SL-T networks SL-T$_{1}$ and SL-T$_{2}$ are exposed to the supervision information of the teacher network to obtain the teacher knowledge. At the same time, the outputs of SL-T$_{1}$ and SL-T$_{2}$ are weighted fused, and the fused output is provided to the student network, allowing it to distill learning ability under this supervision information. }
\label{pic:pic4}
\end{figure}
\section{Method}
\subsection{Background}
The output of a deep neural network reflects its confidence for all categories. In the vanilla KD, the student network strives to approximate the teacher network via the metric criterion $m(a,b)$, where $a$ and $b$ represent the output of the student and teacher, respectively. A pre-trained teacher network performs better, and its output is close to the ground truth. Therefore, making $m(a,b)=0$ implies that the student needs approximate the teacher output at each training epoch. When exists a large capacity gap between the student and teacher, the distillation process will become challenging, as the teacher output is too precise, while the student has limited learning ability, and the value of $m(a,b)$ is too large, which ultimately leads to unsatisfactory knowledge distillation performance. Therefore, we suppose that the vanilla KD neglects the learning ability of the teacher, and exclusively learning from the teacher may be challenging for the student. Given these insights, we introduce a novel concept incorporating the teacher network learning ability into the vanilla KD as an auxiliary knowledge transfer mechanism. However, the learning ability of a neural network is challenging to measure. We empirically observed, inspired by \citep{zhang2022quantifying} and the information bottleneck theory\citep{shwartz2017opening,wolchover2017new}, that in the procedure of knowledge distillation, different student networks exhibit distinct learning abilities and therefore display dissimilar learning trajectories. Thus, the diversity of learning abilities is manifested through dissimilar learning trajectories. In other words, if a student network is supervised by the learning trajectory of a teacher network, they can acquire the teacher learning ability. However, to the best of our knowledge, there are no direct means to access the teacher learning trajectory.
\subsection{Transferring Teacher Learning Ability}
\label{Overall}
To circumvent the above restriction, we introduce a self-learning teacher(SL-T) network with the same architecture as the teacher but without pre-training. Supervised by the learning trajectory of SL-T, the student can indirectly obtain the teacher network learning trajectory and thereby access its learning ability. Specifically, during the training epoch $i$, the student network $S$ is exposed to training inputs $(X,Y)$, and is trained under the supervision information provided by the teacher $T$:
\begin{equation}
\begin{aligned}
    L_{TS}(\phi(i))=\alpha\mathrm{CE}(Y,S(X;\phi(i)))&+(1-\alpha)\mathrm{KD}(T(X;\theta)/\tau,S(X;\phi(i))/\tau)\\
    \phi(i+1)&\leftarrow\underset{\phi}{min}L_{TS}(\phi(i))
\end{aligned} 
\label{eq:eq1}
\end{equation}
where $\phi(i)$ and $\theta$ represent the parameters of the student network $S$ and teacher network $T$ at epoch $i$,  $\mathrm{CE}$ refers to the cross-entropy loss function, $\mathrm{KD}$ measures the difference between the output of the student and teacher, we use KL-Divergence in our work, $\alpha$ is the coefficient that controls these two losses of weights, $\tau$ is the hyper-parameter temperature to soften the output. At the same time, treating the SL-T network $SLT$ as a student and exposing them to the supervision information from the teacher $T$:
\begin{equation}
\begin{split}
L_{TSLT}(\beta(i))=\alpha\mathrm{CE}(Y,SLT(X;\beta(i)))&+(1-\alpha)\mathrm{KD}(T(X;\theta)/\tau,SLT(X;\beta(i))/\tau)\\
\beta(i+1)&\leftarrow\underset{\beta}{min}L_{TSLT}(\beta(i))
\label{eq:eq2}
\end{split} 
\end{equation}
Through the above steps, we hope that the student and SL-T network will be exposed to teacher supervision information, and their different network architecture will lead to distinct learning trajectories. The learning trajectory of the SL-T network can be approximately equivalent to that of the teacher. Thus we enable the student $S$ to learn under the supervision information of the SL-T network $SLT$:
\begin{equation}
\begin{split}
L_{SLTS}(\phi(i))=\alpha\mathrm{CE}(Y,S(X;\phi(i)))&+(1-\alpha)\mathrm{KD}(SLT(X;\beta(i))/\tau,S(X;\phi(i))/\tau)\\
\phi(i+1)&\leftarrow\underset{\phi}{min}L_{SLTS}(\phi(i))
\label{eq:eq3}
\end{split} 
\end{equation}\\
Unlike the teacher, the SL-T network constantly learns throughout the knowledge distillation. Their output is far from the teacher output while closer to the student output at the initial stage(as shown in Figure\ref{pic:pic2}). Therefore, it is easier to make $m(a,b^{sl})=0$ than to make $m(a,b)=0$, where $b^{sl}$ is the SL-T network output. Moreover, after the knowledge distillation process is completed, the student acquires the learning ability of the teacher by tracing the learning trajectory of the SL-T and thus learns in a more teacher-like manner. The overall framework of our proposed method as shown in Figure\ref{pic:pic4}, the total loss for the student network is as follows:
\begin{equation}
\begin{split}
 L_{Total}(\phi(i))&=\lambda L_{TS}(\phi(i))+\eta L_{SLTS}(\phi(i))\\
 \phi(i+1)&\leftarrow\underset{\phi}{min}L_{Total}(\phi(i))
 \label{eq:eq7}
\end{split}
\end{equation}
where $\lambda$ and  $\eta$ are coefficients that control the weight of supervision information from the teacher and SL-T.
\subsection{Learning Trajectory Enhancement}
\begin{figure}[h]
\centering
\begin{tabular}{cc}
{\includegraphics[width=5cm]{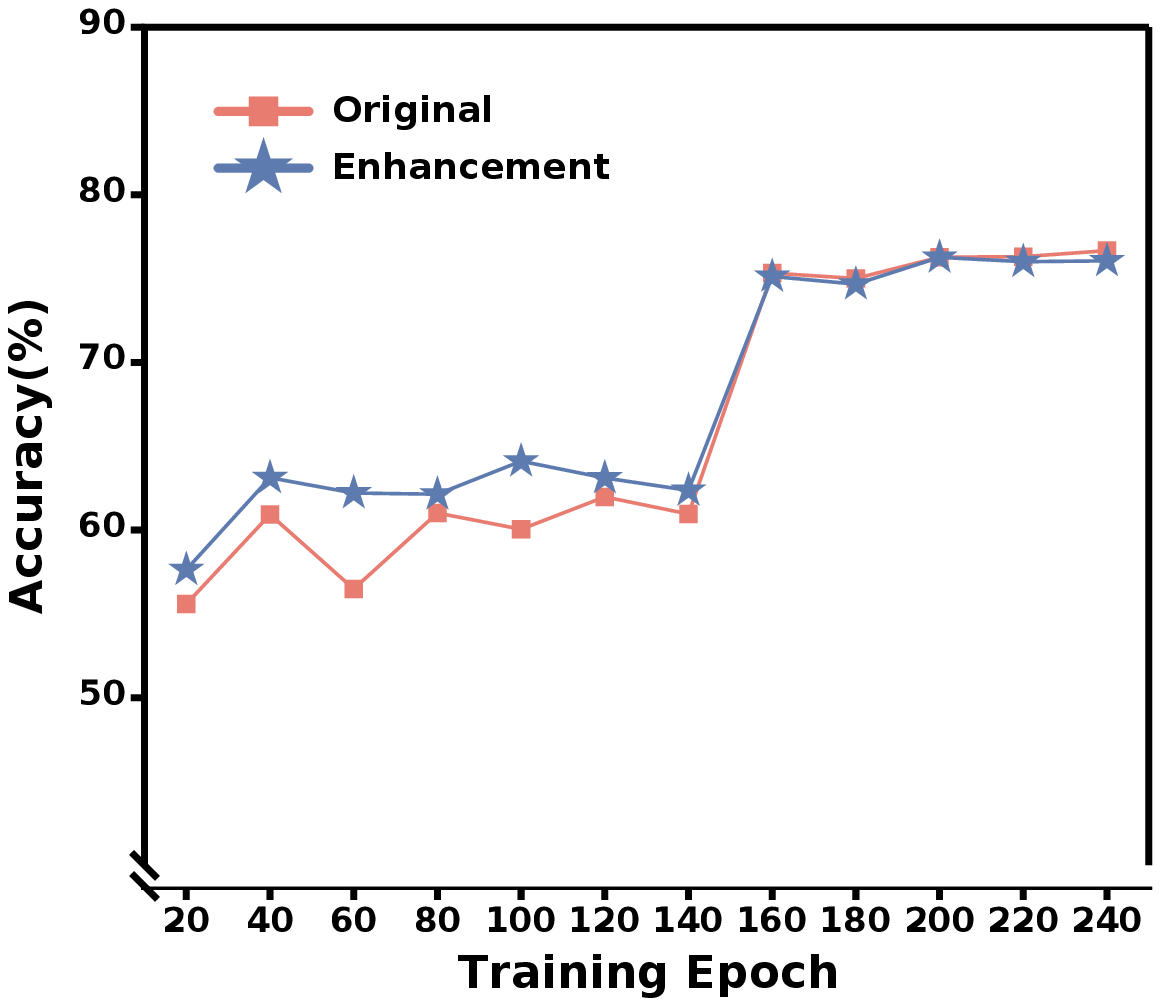}}&{\includegraphics[width=5cm]{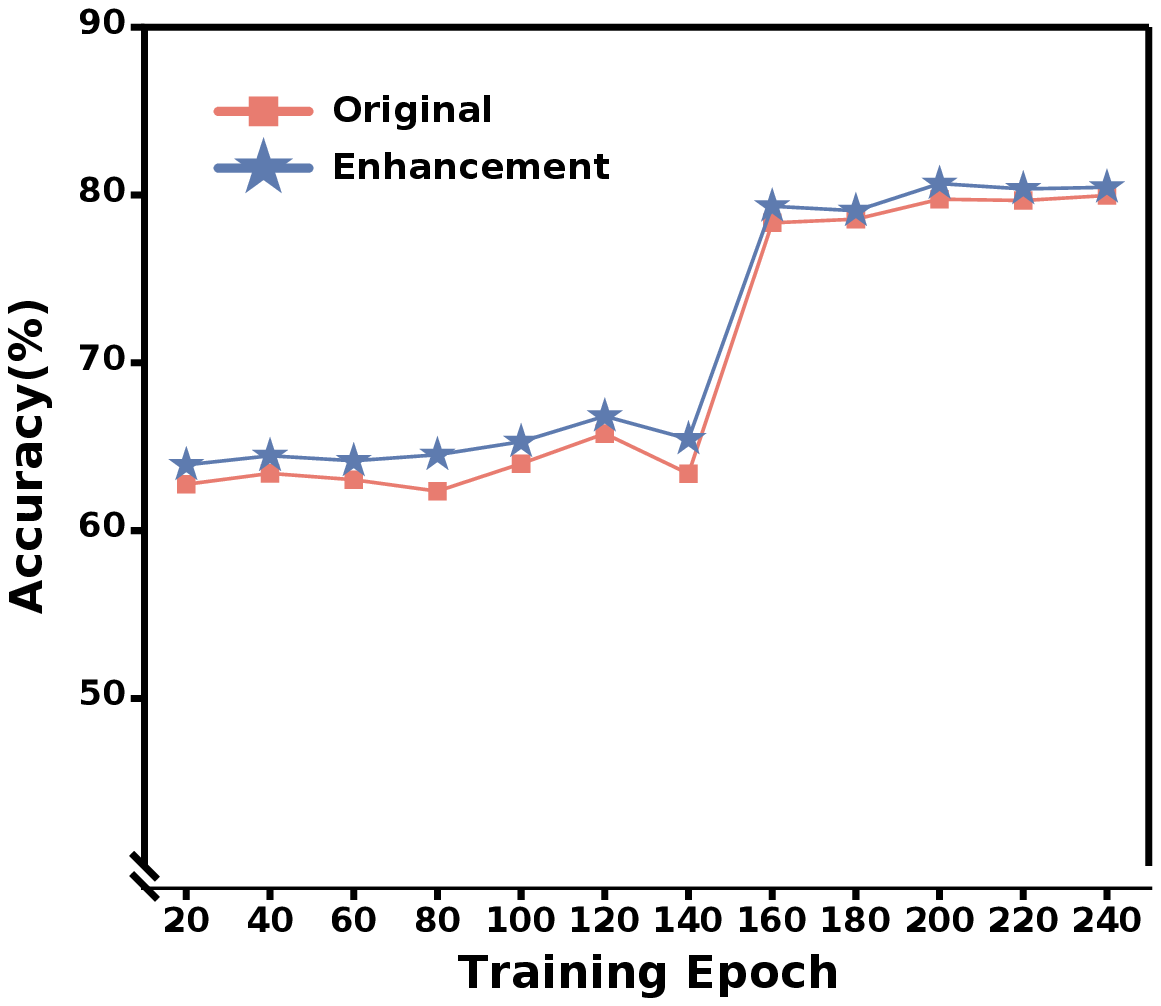}}\\
{(a) VGG13}&{(b) ResNet32x4}\\
\end{tabular}
\caption{The test accuracy(\%) of the SL-T on CIFAR-100. (a) The SL-T is VGG13. (b) The SL-T is ResNet32x4. The symbol "Original" denotes adopting an ordinary way to train SL-T, while "Enhancement" denotes adopting the learning trajectory enhancement strategy.}
\label{pic:pic3}
\end{figure}

In most knowledge distillation methods, the initial step is to typically search among all pre-trained models to identify the optimal teacher model for the distillation process. However, it is not possible to select the optimal SL-T model in a teacher-like manner due to the synchronized learning of the SL-T and student. The method of multi-teacher knowledge distillation integrates the output of multiple teachers as supervision information and has been shown to yield superior distillation results compared to conventional single-teacher methods. Considering the accuracy and effectiveness of the supervision information provided by the SL-T, we adopt the concept of the multi-teacher knowledge distillation\citep{you2017learning,fukuda2017efficient,wu2019multi,du2020agree,kwon2020adaptive} methods and introduce multiple completely identical SL-Ts, which outputs are weighted fused and provided to the student as supervision information. The learning trajectory enhancement strategy can enable SL-Ts to provide the student with more reasonable approximation goals in each iteration. Specifically, in the process of training SL-T, we introduce two fully isomorphic and untrained deep neural networks SL-T$_{1}$($SLT_{1}$) and SL-T$_{2}$($SLT_{2}$), which learn from the teacher as student individually during the entire process of knowledge distillation:
\begin{equation}
\begin{aligned}
L_{TSLT_{1}}(\beta_{1}(i))=\alpha\mathrm{CE}(Y,SLT_{1}(X;\beta_{1}(i)))&+(1-\alpha)\mathrm{KD}(T(X;\theta)/\tau,SLT_{1}(X;\beta_{1}(i))/\tau)\\
\beta_{1}(i+1)&\leftarrow\underset{\beta_{1}}{min}L_{TSLT_{1}}(\beta_{1}(i))
\label{eq:eq4}
\end{aligned} 
\end{equation}
\begin{equation}
\begin{split}
L_{TSLT_{2}}(\beta_{2}(i))=\alpha\mathrm{CE}(Y,SLT_{2}(X;\beta_{2}(i)))&+(1-\alpha)\mathrm{KD}(T(X;\theta)/\tau,SLT_{2}(X;\beta_{2}(i))/\tau)\\
\beta_{2}(i+1)&\leftarrow\underset{\beta_{2}}{min}L_{TSLT_{2}}(\beta_{2}(i))
\label{eq:eq5}
\end{split} 
\end{equation}
where $\beta_{1}(i)$ and $\beta_{2}(i)$ represent the parameter of SL-T$_{1}$ and SL-T$_{2}$ at epoch $i$. We then weighted fuse the output of these two SL-T networks :
\begin{equation}
SLT(X;\beta(i))=\rho SLT_{1}(X;\beta_1(i))+(1-\rho)SLT_{2}(X;\beta_2(i))
\label{eq:eq6}
\end{equation}    
where $\rho$ is the weight coefficient, the fused output is provided to the student network as supervision information(Eq\ref{eq:eq3}). In this way, we hope it can improve the reliability of the learning trajectory provided by SL-T, thereby improving the overall performance of knowledge distillation (as shown in Figure\ref{pic:pic3}).
\begin{table}[t]
\caption{\textbf{Results on the CIFAR-100 validation set.} Teachers and students use the same architectures. The symbol ↑ represents the performance improvement over the vanilla KD. All results are the average over 5 trials.}
\label{tab:tab1}
\centering
\begin{tabular}{ccccccc}
\toprule[2pt]
\multirow{2}{*}{Teacher}&{ResNet32x4}&{ResNet56}&{WRN\_40\_2}&{WRN\_40\_2}&{VGG13}&{VGG13}\\
{}&{79.42}&{72.34}&{75.61}&{75.61}&{74.64}&{74.64}\\
\multirow{2}{*}{Student}&{ResNet8x4}&{ResNet8}&{WRN\_40\_1}&{WRN\_16\_2}&{VGG11}&{VGG8}\\
{}&{72.50}&{59.98}&{71.98}&{73.26}&{71.32}&{70.36}\\
\midrule[1pt]
{AT\citep{zagoruyko2016paying}}&{73.44}&{58.41}&{72.77}&{74.08}&{73.63}&{71.43}\\
{SP\citep{tung2019similarity}}&{72.94}&{58.49}&{72.65}&{73.89}&{74.88}&{73.36}\\
{RKD\citep{park2019relational}}&{71.90}&{58.06}&{72.22}&{73.35}&{73.29}&{71.48}\\
{OFD\citep{heo2019comprehensive}}&{74.95}&{57.42}&{74.33}&{75.24}&{75.14}&{73.95}\\
{CRD\citep{tian2019contrastive}}&{75.51}&{60.94}&{74.14}&{75.48}&{74.02}&{73.94}\\
{ReviewKD\citep{chen2021distilling}}&{75.63}&{60.27}&{\textbf{75.09}}&{76.12}&{75.32}&{74.84}\\
{DKD\citep{zhao2022decoupled}}&{76.32}&{61.62}&{74.81}&{76.24}&{74.92}&{74.68}\\
{KD\citep{hinton2015distilling}}&{73.33}&{60.17}&{73.54}&{74.92}&{74.54}&{72.98}\\
{SLKD}&{\textbf{76.85}}&{\textbf{62.05}}&{74.91}&{\textbf{76.31}}&{\textbf{75.46}}&{\textbf{75.01}}\\
{$\uparrow$}&{+3.52}&{+1.88}&{+1.37}&{+1.39}&{+0.92}&{+2.03}\\
\bottomrule[2pt]
\end{tabular}
\end{table}
\section{Experiment}
In this section, we evaluate our method on various popular neural networks, e.g., VGG\citep{simonyan2014very}, ResNet\citep{he2016deep}, WRN\citep{zagoruyko2016wide}, MobileNetV1\citep{howard2017mobilenets}, MobileNetV2\citep{sandler2018mobilenetv2}, ShuffleNetV1\citep{zhang2018shufflenet} and ShuffleNetV2\citep{ma2018shufflenet}. On image classification, we leverage the commonly used benchmark CIFAR-100\citep{krizhevsky2009learning} and ImageNet\citep{deng2009imagenet} to evaluate the effectiveness of the proposed method. Our approach achieves superior performance compared to many state-of-the-art methods.
\subsection{Experimental Settings}
\paragraph{Dataset.}
The CIFAR-100\citep{krizhevsky2009learning} is a well-known image classification dataset containing 32×32 images of 100 categories. Training and validation sets are composed of 50k and 10k images. We report top-1 accuracy as the evaluation metric. ImageNet\citep{deng2009imagenet} is a large-scale image classification dataset that contains more than one million training samples with 1000 categories. We report the top-1 and top-5 accuracies to measure the model performances.
\begin{table}[t]
\caption{\textbf{Results on the CIFAR-100 validation set.} Teachers and students use different architectures. The symbol ↑ represents the performance improvement over the vanilla KD. All results are the average over 5 trials.}
\label{tab:tab2}
\centering
\begin{tabular}{cccccc}
\toprule[2pt]
\multirow{2}{*}{Teacher}&{ResNet32x4}&{ResNet32x4}&{ResNet50}&{VGG13}&{WRN\_40\_2}\\
{}&{79.42}&{79.42}&{79.34}&{74.64}&{75.61}\\
\multirow{2}{*}{Student}&{ShuffleNetV2}&{ShuffleNetV1}&{MobileNetV2}&{MobileNetV2}&{ShuffleNetV1}\\
{}&{71.82}&{70.50}&{64.60}&{64.60}&{70.50}\\
\midrule[1pt]
{AT\citep{zagoruyko2016paying}}&{72.73}&{71.73}&{58.58}&{59.40}&{73.32}\\
{SP\citep{tung2019similarity}}&{75.75}&{75.51}&{68.61}&{66.74}&{75.94}\\
{RKD\citep{park2019relational}}&{73.21}&{72.28}&{64.43}&{64.52}&{72.21}\\
{OFD\citep{heo2019comprehensive}}&{76.82}&{75.98}&{69.04}&{69.48}&{75.85}\\
{CRD\citep{tian2019contrastive}}&{75.65}&{75.11}&{69.11}&{69.73}&{76.05}\\
{ReviewKD\citep{chen2021distilling}}&{\textbf{77.78}}&{\textbf{77.45}}&{69.89}&{\textbf{70.37}}&{\textbf{77.14}}\\
{DKD\citep{zhao2022decoupled}}&{77.07}&{76.45}&{70.35}&{69.71}&{76.70}\\
{KD\citep{hinton2015distilling}}&{74.45}&{74.07}&{67.35}&{67.37}&{74.83}\\
{SLKD}&{77.26}&{76.17}&{\textbf{70.43}}&{69.94}&{76.72}\\
{$\uparrow$}&{+2.81}&{+2.10}&{+3.08}&{+2.57}&{+1.89}\\
\bottomrule[2pt]
\end{tabular}
\end{table}
\paragraph{Training details.}
We follow the implementation details of previous works\citep{zhao2022decoupled,chen2021distilling} and report the experimental results of all teacher-student pairs. Specifically, we use SGD optimizer\citep{sutskever2013importance} for all datasets. For CIFAR-100, the total training epoch is set to 240. The initial learning rate is 0.05 for VGG, ResNet, and WRN, 0.01 for ShuffleNetV1, ShuffleNetV2, and MobileNetV2, with a decay rate of 0.1 at epoch 150, 180, and 210. The batch size is 64, and the weight decay is 5$\times$$10^{-4}$. For ImageNet, The initial learning rate is 0.2 and decays by 0.1 at epoch 30, 60, and 90. The batch size is 512, and the weight decay is  1$\times$$10^{-4}$. More details are attached in the supplement.
\begin{table}[h]
\caption{\textbf{Top-1 and top-5 accuracy(\%) on the ImageNet validation.} We set ResNet-34 as the teacher and ResNet-18 as the student. All results are the average over 3 trials.}
\label{tab:tab3}
\centering
\resizebox{\textwidth}{!}{
\begin{tabular}{cccccccccc}
\toprule[2pt]
{}&{Teacher}&{Student}&{AT\citep{zagoruyko2016paying}}&{CRD\citep{tian2019contrastive}}&{OFD\citep{heo2019comprehensive}}&{ReviewKD\citep{chen2021distilling}}&{DKD\citep{zhao2022decoupled}}&{KD\citep{hinton2015distilling}}&{SLKD}\\
\midrule[1pt]
{Top-1}&{73.31}&{69.75}&{70.69}&{71.17}&{70.81}&{71.61}&{71.70}&{70.66}&{\textbf{71.82}}\\
{Top-5}&{91.42}&{89.07}&{90.01}&{90.13}&{89.98}&{90.51}&{90.41}&{89.88}&{\textbf{90.56}}\\
\bottomrule[2pt]
\end{tabular}
}
\end{table}
\subsection{Main Results}
\paragraph{Results on CIFAR-100.} We discuss the experimental results on CIFAR-100 to show the generalization ability of our SLKD. The validation accuracy is reported in Table \ref{tab:tab1} and Table \ref{tab:tab2}. Table \ref{tab:tab1} contains the results where teachers and students are of the same network architecture. Table \ref{tab:tab2} shows the results where teachers and students are of different series.\\
As shown in Table\ref{tab:tab1} and \ref{tab:tab2}, SLKD consistently improves all teacher-student pairs compared with the baseline and the vanilla KD. Our method enhances 1$\sim$3$\%$ and 2$\sim$3$\%$ on teacher-student pairs of the same and different series, evidencing the effectiveness and generalization ability of our method. The results when distilling VGG11 and VGG8 from VGG13 are interesting. In this case, the validation accuracy of students even surpasses the teacher, further demonstrating the superiority of SLKD.
\paragraph{Results on ImageNet.} We further evaluate our method on a challenging dataset. Here, we report the result where the teacher and student are of the same network architecture in Table \ref{tab:tab3}, and the result where the teacher and student are from different series in Table \ref{tab:tab4}. We show the top-1 and top-5 accuracies to verify the effectiveness of the proposed method. Our SLKD outperforms the vanilla KD by a significant margin and is better than most feature-based distillation methods.
\begin{table}[h]
\caption{\textbf{Top-1 and top-5 accuracy(\%) on the ImageNet validation.} We set ResNet-50 as the teacher and MobileNet-V1 as the student. All results are the average over 3 trials.}
\label{tab:tab4}
\centering
\resizebox{\textwidth}{!}{
\begin{tabular}{cccccccccc}
\toprule[2pt]
{}&{Teacher}&{Student}&{AT\citep{zagoruyko2016paying}}&{CRD\citep{tian2019contrastive}}&{OFD\citep{heo2019comprehensive}}&{ReviewKD\citep{chen2021distilling}}&{DKD\citep{zhao2022decoupled}}&{KD\citep{hinton2015distilling}}&{SLKD}\\
\midrule[1pt]
{Top-1}&{76.16}&{68.87}&{69.56}&{71.37}&{71.25}&{\textbf{72.56}}&{72.05}&{68.58}&{72.42}\\
{Top-5}&{92.86}&{88.76}&{89.33}&{90.41}&{90.34}&{91.00}&{91.05}&{88.98}&{\textbf{91.11}}\\
\bottomrule[2pt]
\end{tabular}
}
\end{table}
\paragraph{Improving distillation performances of big teachers.} In the procedure of knowledge distillation, the larger network can not always as a good teacher\citep{lopez2015unifying,mirzadeh2020improved}. Specifically, a more powerful or larger teacher is not necessary to transfer more beneficial knowledge, even resulting inferior performances than smaller ones. Previous works\citep{cho2019efficacy, wang2021knowledge} attributed this phenomenon to the large capacity gap between big teachers and small students. We suppose that the main reason for this problem is that students do not have the same learning ability as teachers, and the output of teachers is too precise, which causes unsatisfactory distillation performance. Some related works about this problem also could be used to explain this problem, e.g., ESKD\citep{cho2019efficacy} alleviates this problem by employing early-stopped teacher models, and these teachers would be under-optimal. Pro-KD\citep{rezagholizadeh2021pro} defines a smoother training path for the student network by following the training footprints of the teacher instead of relying on distilling from a well-trained teacher network. To validate our conjecture, we conduct experiments on a series of teacher models. Experimental results in Table\ref{tab:tab5} and Table\ref{tab:tab6} consistently indicate that our approach alleviates the capacity-gap problem.
\begin{table}[h]
\caption{\textbf{Results on CIFAR-100.} We set ResNet20 as the student and ResNet series networks as teachers.}
\label{tab:tab5}
\centering
\begin{tabular}{ccccc}
\toprule[2pt]
\multirow{2}{*}{Teacher}&{ResNet56}&{ResNet110}&{ResNet50}&{ResNet32x4}\\
{}&{72.34}&{74.31}&{79.34}&{79.42}\\
\midrule[1pt]
{KD}&{71.04}&{70.71}&{70.55}&{69.21}\\
{SLKD}&{71.13}&{71.20}&{70.98}&{70.38}\\
\bottomrule[2pt]
\end{tabular}
\end{table}
\begin{table}[h]
\caption{\textbf{Results on CIFAR-100.} We set WRN\_16\_2 as the student and networks from different series as teachers.}
\label{tab:tab6}
\centering
\begin{tabular}{ccccc}
\toprule[2pt]
\multirow{2}{*}{Teacher}&{ResNet110}&{VGG13}&{WRN\_40\_2}&{ReNet32x4}\\
{}&{74.31}&{74.64}&{75.61}&{79.42}\\
\midrule[1pt]
{KD}&{74.94}&{74.69}&{75.31}&{74.37}\\
{SLKD}&{75.29}&{75.42}&{76.31}&{75.08}\\
\bottomrule[2pt]
\end{tabular}
\end{table}
\subsection{Ablation Study}
\paragraph{Training students with SL-T only.}
The proposed SLKD enables the student network to learn from the teacher and SL-T network. In order to validate the effectiveness of the supervision information provided by the SL-T, we conducted experiments in which the student was exposed to the supervision information provided solely by either the teacher or the SL-T. The results in Table \ref{tab:tab7} verify that, compared to the teacher, the supervision information provided by the SL-T is more conducive to student learning, resulting in better distillation results. Combining them could further boost performance, confirming our initial idea that learning ability can be seen as knowledge to transfer.
\paragraph{Parallel vs. sequential training.}
In section\ref{Overall}, we enable the student to receive supervision information in parallel from both the teacher and SL-T networks. Alternatively, a sequential approach can be employed, where the SL-T first learns from the teacher. Subsequently, the student is exposed to the supervision information provided by the teacher and the SL-T to distill knowledge and learning ability. We conducted comparative experiments under the same experimental settings to show the differences between these two learning approaches. The results are reported in Table \ref{tab:tab8} and Table \ref{tab:tab9}.
\begin{table}[h]
\caption{\textbf{Results on CIFAR-100.} We compare the performance of the student under the supervision information provided by Teacher(T) and SL-T respectively.}
\label{tab:tab7}
\centering
\begin{tabular}{ccccc}
\toprule[2pt]
{Teacher}&{Student}&{T}&{SL-T}&{Accuracy}\\
\midrule[1pt]
\multirow{3}{*}{ResNet32x4}&\multirow{3}{*}{ResNet8x4}&{\ding{52}}&{\ding{56}}&{73.33}\\
{}&{}&{\ding{56}}&{\ding{52}}&{75.71}\\
{}&{}&{\ding{52}}&{\ding{52}}&{76.13}\\
\multirow{3}{*}{VGG13}&\multirow{3}{*}{MobileNetV2}&{\ding{52}}&{\ding{56}}&{67.37}\\
{}&{}&{\ding{56}}&{\ding{52}}&{68.76}\\
{}&{}&{\ding{52}}&{\ding{52}}&{69.25}\\
\bottomrule[2pt]
\end{tabular}
\end{table}
\begin{table}[h]
\caption{\textbf{CIFAR-100 validation accuracy.} Teachers and students use the same architectures. SLKD* means adopting the sequential learning approach.}
\label{tab:tab8}
\centering
\begin{tabular}{lcccccc}
\toprule[2pt]
\multirow{2}{*}{Teacher}&{ResNet32x4}&{ResNet56}&{WRN\_40\_2}&{WRN\_40\_2}&{VGG13}&{VGG13}\\
{}&{79.42}&{72.34}&{75.61}&{75.61}&{74.64}&{74.64}\\
\multirow{2}{*}{Student}&{ResNet8x4}&{ResNet8}&{WRN\_40\_1}&{WRN\_16\_2}&{VGG11}&{VGG8}\\
{}&{72.50}&{59.98}&{71.98}&{73.26}&{71.32}&{70.36}\\
\midrule[1pt]
{SLKD}&{76.85}&{62.05}&{74.91}&{76.31}&{75.46}&{75.01}\\
{SLKD*}&{75.61}&{61.99}&{74.60}&{75.78}&{74.96}&{74.41}\\
\bottomrule[2pt]
\end{tabular}
\end{table}
\begin{table}[h]
\caption{\textbf{CIFAR-100 validation accuracy.} Teachers and students use different architectures. SLKD* means adopting the sequential learning approach.}
\label{tab:tab9}
\centering
\begin{tabular}{lcccccc}
\toprule[2pt]
\multirow{2}{*}{Teacher}&{ResNet32x4}&{ResNet32x4}&{ResNet50}&{VGG13}&{WRN\_40\_2}\\
{}&{79.42}&{79.42}&{79.34}&{74.64}&{75.61}\\
\multirow{2}{*}{Student}&{ShuffleNetV2}&{ShuffleNetV1}&{MobileNetV2}&{MobileNetV2}&{ShuffleNetV1}\\
{}&{71.82}&{70.50}&{64.60}&{64.60}&{70.50}\\
\midrule[1pt]
{SLKD}&{77.26}&{76.17}&{70.43}&{69.94}&{76.72}\\
{SLKD*}&{76.12}&{75.34}&{68.29}&{69.36}&{76.29}\\
\bottomrule[2pt]
\end{tabular}
\end{table}
\section{Conclusion and Discussion}
This work develops a framework for knowledge distillation through transferring learning ability that enables the student to learn more teacher-likely. Through theoretical analysis, we reveal that most existing methods neglect the importance of learning ability. Therefore, distillation performance may be inferior to training from scratch when the capacity gap is too large. This paper introduced an SL-T network to transfer the teacher network learning ability. A learning trajectory enhancement strategy is designed to ensure the accuracy and effectiveness of the supervision information provided by the SL-T. The proposed SLKD achieves significant improvements on CIFAR-100 and ImageNet datasets for image classification tasks. We hope this study will contribute to future knowledge distillation research.
\paragraph{Limitation and future works.}
An SL-T network is introduced to transfer the teacher network learning ability, introducing extra computational and storage costs. How to develop a computation-free and storage-free alternative needs future investigation. Another area for improvement is that our technique is a logits-based method and more suitable for image classification tasks. Developing a feature-based variant of our technique for other tasks that greatly depends on the quality of deep features is also worthwhile.
\small
\bibliographystyle{unsrt}
\bibliography{ref}
\newpage
\begin{appendices}
\section{Appendix}
\subsection{Computing Infrastructure}
All of the experiments are conducted with PyTorch. CIFAR-100 experiments are conducted on a server containing four NVIDIA GeForce RTX 3090 GPUs with 24GB RAM. The CUDA version is 11.6. ImageNet experiments are conducted on a server containing eight NVIDIA GeForce RTX 4090 GPUs with 24GB RAM. The CUDA version is 12.0.
\subsection{Implementation of SLKD}
We present the pseudo code of SLKD in this section, as shown in Algorithm\ref{alg1}. The training code for the experiments is available at \url{https://github.com/BrilliantCheng/SLKD}.
\begin{algorithm}

\caption{Pseudo code of SLKD in a Pytorch-like style}
\label{alg1}
\LinesNumbered
\vspace{5mm}
\# l\_tea: teacher output logits\\
\# l\_stu: student output logits\\
\# l\_slt1: self-learning teacher1 output logits\\
\# l\_slt2: self-learning teacher2 output logits\\
\# $\tau$: temperature for SLKD\\
\# y: ground truth\\
\# $\alpha$, $\lambda$, $\eta$, $\rho$: hyper-parameters for SLKD\\
\vspace{5mm}
\# teacher \& student\\
L\_TS = $\alpha$ $\cdot$ F.cross\_entropy(l\_stu, y) + (1 - $\alpha$) $\cdot$ kd\_loss(l\_stu, l\_tea, $\tau$)\\
\vspace{5mm}
\# teacher \& self-learning teacher\\
L\_TSLT1 = $\alpha$ $\cdot$ F.cross\_entropy(l\_slt1, y) + (1 - $\alpha$) $\cdot$ kd\_loss(l\_slt1, l\_tea, $\tau$)\\
L\_TSLT2 = $\alpha$ $\cdot$ F.cross\_entropy(l\_slt2, y) + (1 - $\alpha$) $\cdot$ kd\_loss(l\_slt2, l\_tea, $\tau$)\\
l\_slt = $\rho$ $\cdot$ l\_slt1 + (1 - $\rho$) $\cdot$ l\_slt2\\
\vspace{5mm}
\# self-learning teacher \& student\\
L\_SLTS = $\alpha$ $\cdot$ F.cross\_entropy(l\_stu, y) + (1 - $\alpha$) $\cdot$ kd\_loss(l\_stu, l\_slt, $\tau$)\\
\vspace{5mm}
\# student total loss\\
L\_TOTAL = $\lambda$ $\cdot$ L\_TS + $\eta$ $\cdot$ L\_SLTS
\end{algorithm}
\subsection{Datasets and Training Details}
We adopt two datasets including CIFAR-100\citep{krizhevsky2009learning} and ImageNet\citep{deng2009imagenet} to conduct experiments. We apply channel-wise mean and standard deviation normalization for all images. A horizontal flip is used for data augmentation. For CIFAR-100, each training image is padded by 4 pixels on each size and randomly cropped as a 32 $\times$ 32 sample. For ImageNet, each image is randomly cropped as a 224 $\times$ 224 sample without padding. We summarize the training hyper-parameters in Table\ref{tab:tab10} and Table\ref{tab:tab11}.
\begin{table}[h]
\caption{\textbf{Training hyper-parameters on CIFAR-100.} BS: bach size; LR: learning rate; WD: weight decay.}
\label{tab:tab10}
\centering
\resizebox{\textwidth}{!}
{
\begin{tabular}{cccccccccc}
\toprule[2pt]
{Student Series}&{Epochs}&{BS}&{LR}&{LR Decay Stages}&{LR Decay Rate}&{Optimizer}&{WD}&{$\alpha$}&{$\tau$}\\
\midrule[1pt]
{VGG, ResNet, WRN}&{240}&{64}&{$5\times10^{-2}$}&{[150, 180, 210]}&{0.1}&{SGD}&{$5\times10^{-4}$}&{0.1}&{4}\\
{ShuffleNetV1, ShuffleNetV2, MobileNetV2}&{240}&{64}&{$1\times10^{-2}$}&{[150, 180, 210]}&{0.1}&{SGD}&{$5\times10^{-4}$}&{0.1}&{4}\\
\bottomrule[2pt]
\end{tabular}
}
\end{table}

\begin{table}[h]
\caption{\textbf{Training hyper-parameters on ImageNet.} BS: bach size; LR: learning rate; WD: weight decay.}
\label{tab:tab11}
\centering
\resizebox{\textwidth}{!}
{
\begin{tabular}{cccccccccc}
\toprule[2pt]
{Student Series}&{Epochs}&{BS}&{LR}&{LR Decay Stages}&{LR Decay Rate}&{Optimizer}&{WD}&{$\alpha$}&{$\tau$}\\
\midrule[1pt]
{ResNet-18, MobileNet-V1}&{100}&{512}&{0.2}&{[30, 60, 90]}&{0.1}&{SGD}&{$1\times10^{-4}$}&{0.5}&{1}\\
\bottomrule[2pt]
\end{tabular}
}
\end{table}

\subsection{Societal Impacts}
Validating the effectiveness of the proposed approach would require substantial computational resources, potentially resulting in increased carbon emissions and environmental implications. Nonetheless, the proposed knowledge distillation method has the potential to enhance the performance of lightweight and compact models. Replacing resource-intensive models with more energy-efficient alternatives in production could result in significant energy savings. Therefore, it is crucial to adequately verify the efficacy of SLKD.
\subsection{Visualization}
\begin{figure}[h]
\centering
\begin{tabular}{cc}
{\includegraphics[width=5cm]{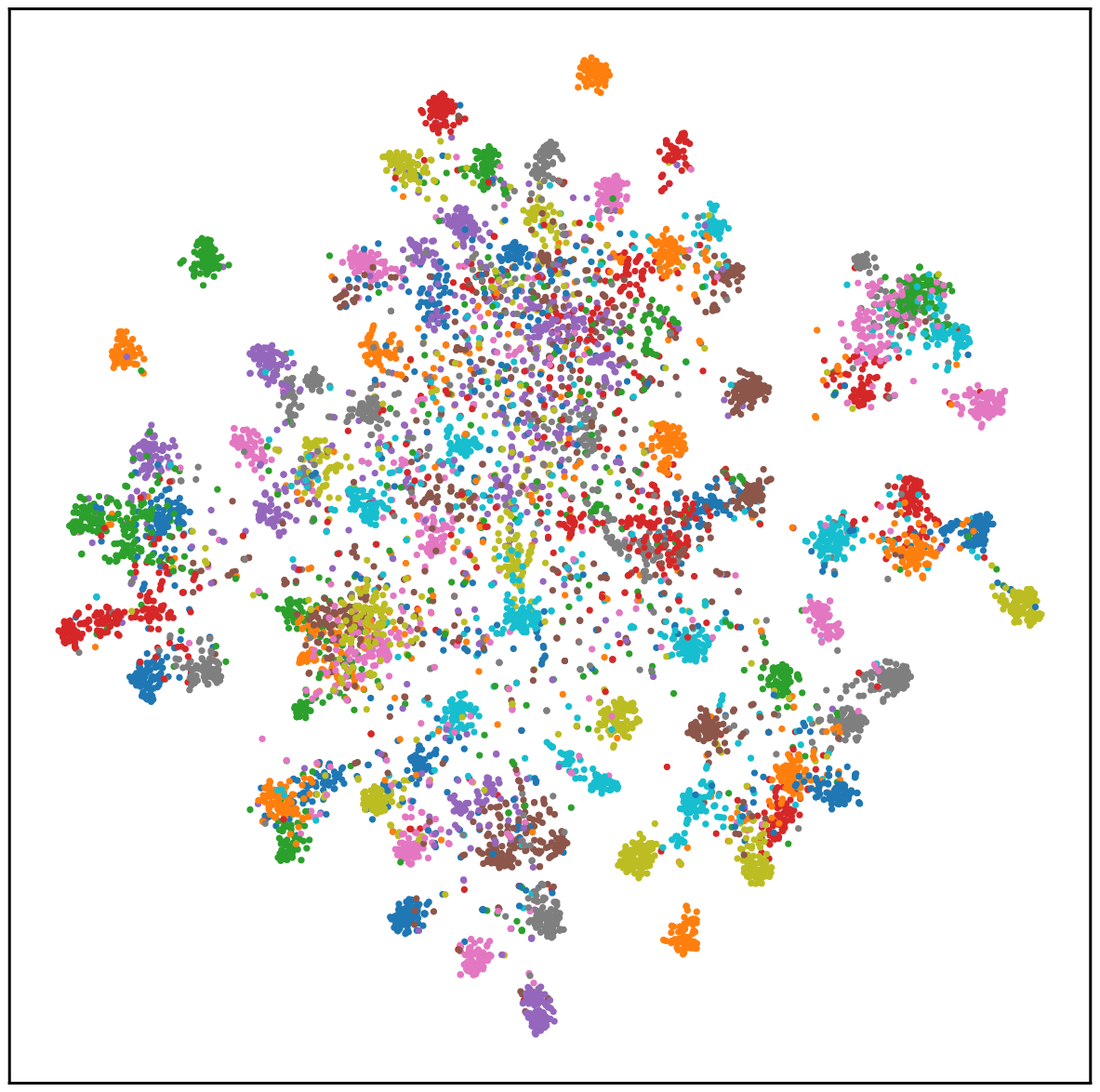}}&{\includegraphics[width=5cm]{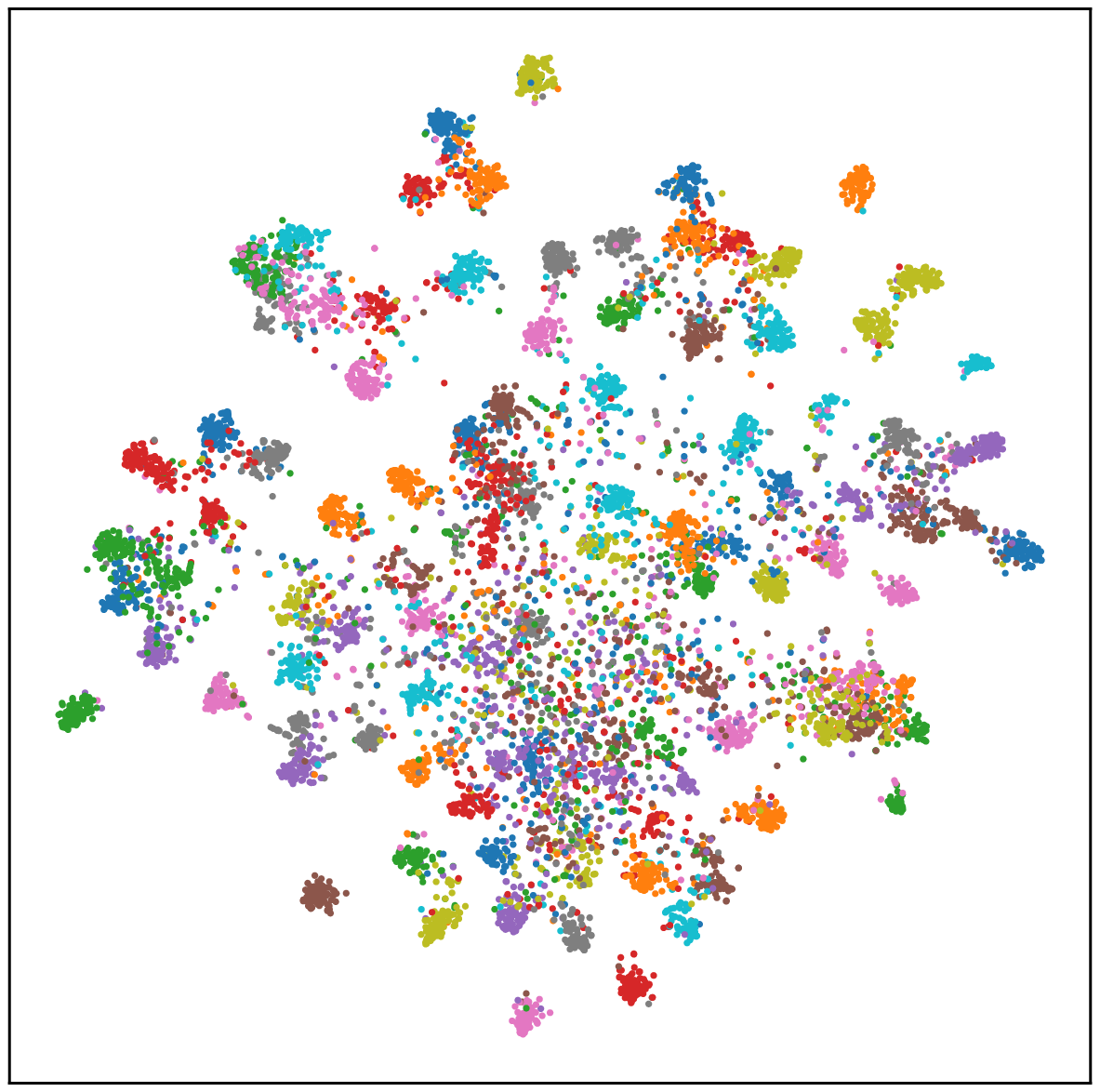}}\\
\end{tabular}
\caption{t-SNE of features learned by KD(left) and SLKD(right).}
\label{pic:pic5}
\end{figure}

\begin{figure}[h]
\centering
\begin{tabular}{cc}
{\includegraphics[width=5cm]{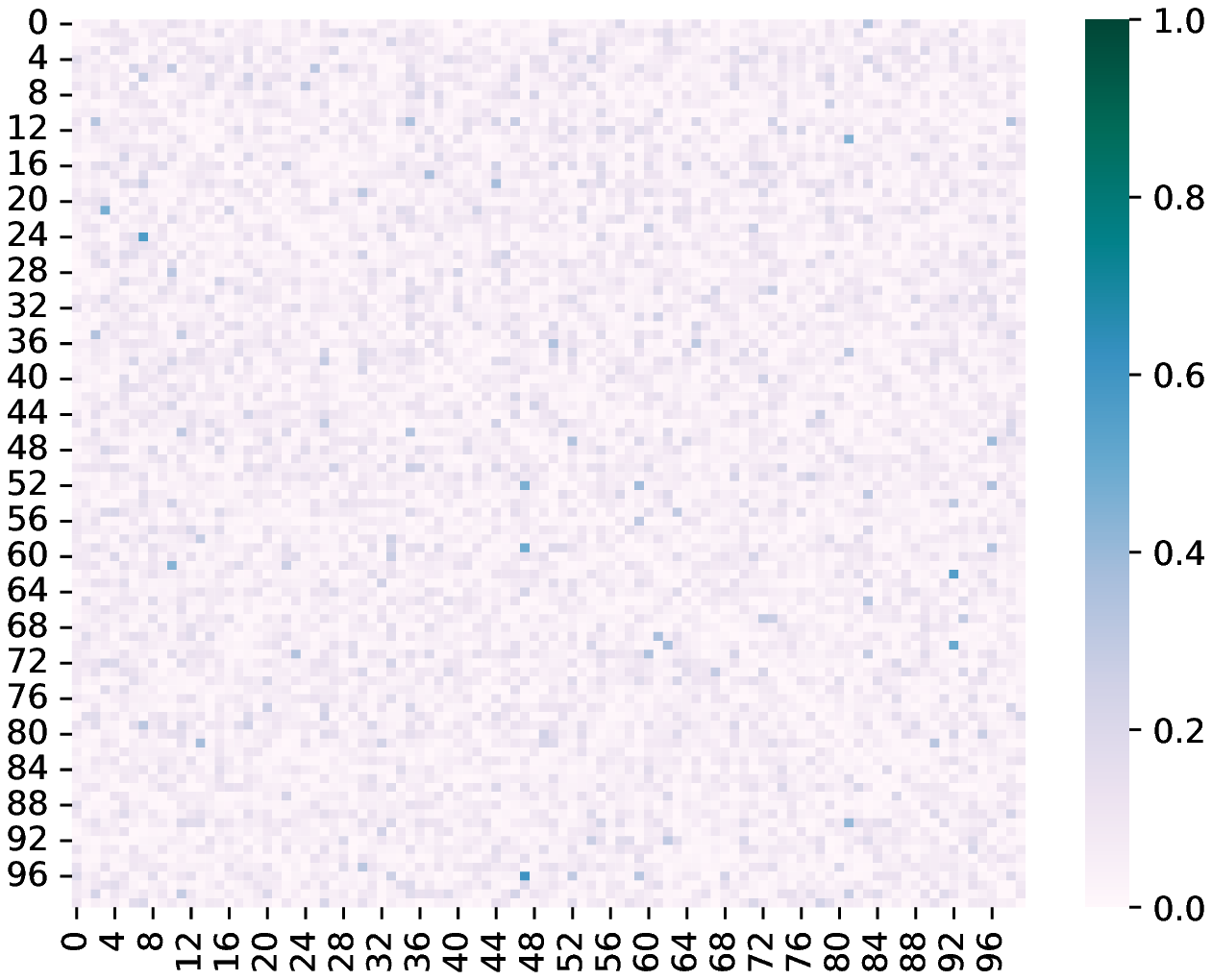}}&{\includegraphics[width=5cm]{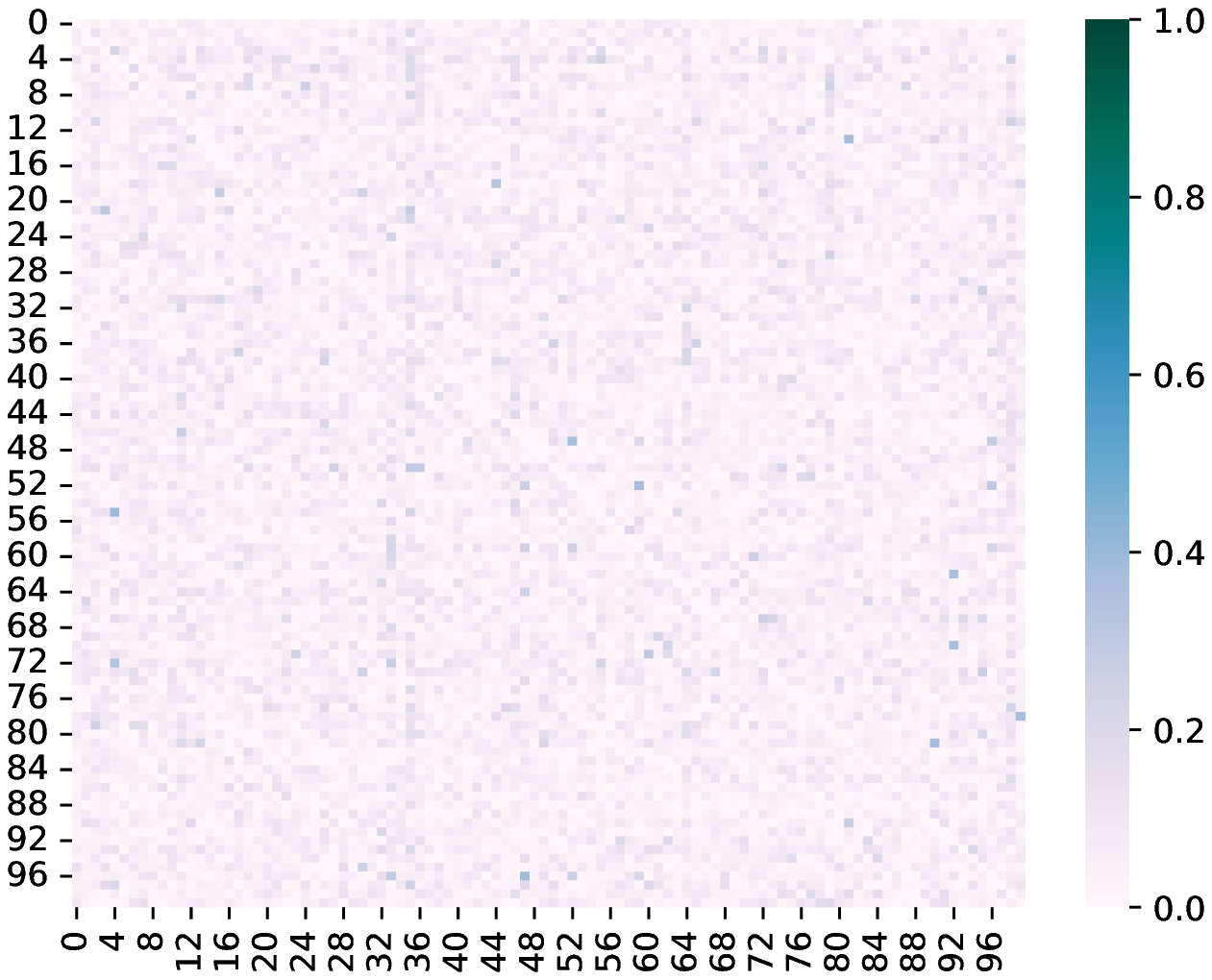}}\\
\end{tabular}
\caption{Difference of correlation matrices of student and teacher output logits. It can be seen that SLKD(right) leads to a smaller difference than KD(left).}
\label{pic:pic6}
\end{figure}

We provide visualizations from two perspectives, using VGG13 as the teacher and VGG8 as the student on CIFAR-100. Firstly, the t-SNE(Figure\ref{pic:pic5}) results demonstrate that the representations generated by SLKD are more separable compared to KD, indicating that our method enhances the discriminability of deep features. Secondly, we visualize the difference between the correlation matrices of student and teacher logits(Figure\ref{pic:pic6}). Our SLKD helps the student to output logits that are more similar to those of the teacher, resulting in better distillation performance compared to KD.

\end{appendices}
\end{document}